\pdfoutput=1
\documentclass{article}

\usepackage[nonatbib,preprint]{neurips_2023}

\usepackage[table,xcdraw]{xcolor} 
\usepackage[utf8]{inputenc} 
\usepackage[T1]{fontenc}    
\usepackage{hyperref}       
\usepackage{url}            
\usepackage{booktabs}       
\usepackage{amsfonts}       
\usepackage{nicefrac}       
\usepackage{microtype}      
\usepackage{amsmath}
\usepackage{graphicx}
\usepackage{lineno}
\usepackage{graphicx}
\usepackage{todonotes}
\usepackage{soul}
\usepackage{siunitx} 
\usepackage{caption}
\usepackage{subcaption}
\usepackage{array}

\usepackage{amsmath}

\title{Operational Wind Speed Forecasts for Chile's Electric Grid Using A Hybrid Machine Learning Model}

\author{%
   Dhruv Suri\thanks{These authors contributed equally to this work. Dhruv Suri's work done while at X - The Moonshot Factory.} \\
  Stanford University\\
  \texttt{surid@stanford.edu} 
  \And
  Praneet Dutta\footnotemark[1] \\
  Google DeepMind\\
  \texttt{praneetdutta@deepmind.com}
  \AND
  Flora Xue \\
  Google DeepMind\\
  \texttt{floraxue@google.com}
  \And
  In\^es Azevedo \\
  Stanford University\\
  \texttt{iazevedo@stanford.edu}
  \And
  Ravi Jain\thanks{Corresponding author.} \\
  X - The Moonshot Factory\\
  \texttt{ravijain@google.com}
}

\begin{document}

\maketitle

\begin{abstract}
As Chile's electric power sector advances toward a future powered by renewable energy, accurate forecasting of renewable generation is essential for managing grid operations. The integration of renewable energy sources is particularly challenging due to the operational difficulties of managing their power generation, which is highly variable compared to fossil fuel sources, delaying the availability of clean energy. To mitigate this, we quantify the impact of increasing intermittent generation from wind and solar on thermal power plants in Chile and introduce a hybrid wind speed forecasting methodology which combines two custom ML models for Chile. The first model is based on TiDE, an MLP-based ML model for short-term forecasts, and the second is based on a graph neural network, GraphCast, for medium-term forecasts up to 10 days. Our hybrid approach outperforms the most accurate operational deterministic systems by 4-21\% for short-term forecasts and 5-23\% for medium-term forecasts and can directly lower the impact of wind generation on thermal ramping, curtailment, and system-level emissions in Chile. 
\end{abstract}

\section{Introduction}

Replacing fossil fuels with renewable energy for electricity generation has a direct impact on climate change, since electricity generation globally contributes to roughly a third of annual carbon emissions \cite{liu2023monitoring}. However, the integration of renewable energy sources into the electric grid is challenging due to the operational difficulties of managing their highly variable power generation. Chile has taken a globally leading role in clean energy and has emerged as a leading destination for solar and wind developers and ambitious clean energy goals for the future \cite{iea_2018}. This makes it a particularly interesting and important area of study for renewable energy forecasting. While Chile's electricity demand is relatively modest (6.5\% of Central and South America's total in 2022), it is growing rapidly, both in aggregate (up from 40 TWh in 2000 to 88 TWh in 2022) and per-capita (70\% over the same period). In 2023, the share of thermal, solar and wind in annual generation in Chile was 38.9\%, 19.9\%, and 11.8\% \cite{infotecnica}.
Electric grid Independent System Operators (ISOs) recognize the need for precise renewable generation and net demand forecasts. Historically, thermal plants provided stable generation with predictable net demand patterns. However, increased penetration of renewables necessitates advanced forecasting due to significant generation ramps from wind and solar. Accurate forecasts are critical for subsequent thermal generator dispatch, which rely on forecast accuracy and security constraints. Among renewable sources, wind power prediction is particularly challenging due to its temporal and spatial variability, and oftentimes lack of sufficient high-quality data, not only for wind power but for wind speed itself \cite{wang2021review}.   
Our work aims to contribute to improving the coordination and control of Chile’s electric power system amidst growing renewable energy integration by enhancing the reliability and accuracy of wind speed forecasts. Our approach diverges from traditional forecast methods by applying machine learning techniques. We introduce a novel hybrid methodology for short and medium-term wind speed forecasts, designed as an input to the ISO’s unit commitment and economic dispatch models in the day-ahead and week-ahead electricity market. For short-term forecasts (0 to 48 hours) we develop a custom ML model specifically optimized for capturing the intricate temporal dependencies in hourly wind speed. For medium-term forecasts (2 to 10 days), we develop a custom ML model based on GraphCast — a graph neural network-based machine learning weather prediction (MLWP) model for medium-range weather forecasts \cite{lam2022graphcast, lam2023learning}. 


\section{Data \& Methods}
\textbf{Data}. We develop a dataset with thermal power plants that includes hourly generation by fuel source obtained from Coordinador Eléctrico Nacional (CEN) \cite{coordinador}, Chile's national ISO. CEN also records historical hourly generation from wind, solar, hydro and geothermal plants, and system demand. For short and medium-term wind magnitude forecasts, we use ERA5, a global reanalysis dataset produced by ECMWF \cite{molteni1996ecmwf}. GraphCast is initially trained on ERA5 data spanning 1979-2016. HRES provides real-time weather predictions using the most recent observations and advanced numerical weather prediction models \cite{rasp2020weatherbench}. HRES-fc0 is a dataset derived from HRES, consisting of the initial state of each high-resolution ensemble forecast and is used as the benchmark for evaluating our hybrid model.

\textbf{Electricity generation marginal response models}. We use a fixed effects ordinary least squares (OLS) regression formulation to model the change in operational behavior of thermal generators in response to a marginal increase in generation from wind and solar. The formulation is a non-logarithmic form of the statistical model used by \cite{bushnell2005ownership} and \cite{suri2024real}.
\begin{equation}
    G_t = \alpha + \beta_1 S_t + \beta_2 W_t + \beta_3 D_t + \beta_4 (W_t - W_{t-1}) + \beta_5 (S_t - S_{t-1}) + \eta_{t,m} + \gamma \textbf{X}_t + \eta_{t,a} + \epsilon
\end{equation}
where $t$ denotes the current timestep; $G_t$, $S_t$, $W_t$ represent the thermal, solar, and wind generation, respectively; $D_t$ is the electricity demand, and the terms between the brackets denote the hourly wind and solar ramp. $\eta_m$ and $\eta_a$ denote the month and year fixed effects. $\textbf{X}_t$ is a set of control variables that includes the hydro and geothermal generation, and imports. The parameters that are of interest are the coefficients for solar, $\beta_1$ and wind, $\beta_2$. These coefficients represent the change in thermal generation in response to a marginal unit increase in generation.

\textbf{Operational wind speed forecast models}. In medium and long-term weather forecasts, the current top deterministic operational system in the world is ECMWF's HRES forecast. Owing to its proven versatility, we use a graph neural network forecasting model, GraphCast, that outperforms the baseline HRES forecasts on 90.3\% of 1380 location targets with an overall skill improvement of 7 to 14\% across global locations \cite{lam2022graphcast}. However, the GraphCast model in \cite{lam2022graphcast} with HRES-fc0 finetuning for Chile proves to be more accurate than HRES forecasts only following a lead time of approximately 7 days. Before this cross-over point, GraphCast’s normalized RMSE relative to HRES ranges from 0 to 1.15 (shown in the SI) in Chile. To overcome this limitation we categorize the forecasting problem into two distinct temporal regimes, short-term and medium-term forecasts. Furthermore, in Chile, day-ahead dispatch processes require advanced hourly generation and net demand forecasts to be made available a day prior for unit commitment and economic dispatch models to be run by the ISO. As GraphCast's time step size is 6 hours, we propose a short-term model with hourly granularity. In order to draw a direct comparison of our methods with ECMWF's HRES, we use a quarter-degree model resolution. However, we focus on those mesh points that are closest to the actual 52 wind farm locations in Chile, and wind speed forecasts are developed for 24 unique quarter-degree interpolated locations in Chile.  


\textbf{Short-term forecasts}. We use an enhanced Time-series Dense Encoder (TiDE) model \cite{das2023long} that represents a state-of-the-art approach for short-term wind speed forecasting from 0 to 48 hours. This model employs a Multi-layer Perceptron (MLP) based encoder-decoder architecture, optimized to handle time-series data with multiple covariates and complex, non-linear dependencies. The encoder processes historical wind speed data and relevant covariates, transforming them into a latent space representation through multiple residual blocks, which are crucial for capturing intricate patterns in the data. The decoder then reconstructs future wind speeds from this latent representation, allowing the model to make accurate predictions up to 48 hours ahead. To enhance the forecast accuracy of TiDE for our application, we integrate a randomized iterative forecasting framework. This integration is inspired by the decomposition-ensemble learning methodology where the data is first decomposed into simpler components, which are then modeled individually using randomized algorithms. 

\textbf{Medium-term forecasts}. For medium-range forecasts, we use GraphCast, a neural network architecture designed to predict weather states by taking the two most recent states of Earth's weather - the current time and six hours earlier - and forecasting the next state six hours ahead. The model uses a 0.25-degree latitude-longitude grid to represent a single weather state, corresponding to roughly 28 km by 28 km resolution at the equator. Implemented as a neural network architecture based on Graph Neural Networks (GNNs) in an "encode-process-decode" configuration, GraphCast consists of 36.7 million parameters. To improve the forecast accuracy of GraphCast, we develop a custom model using a novel multi-stage fine-tuning process. This involved autoregressive fine-tuning on HRES-fc0 data to account for potential distribution shifts between reanalysis data (ERA5) and real-time operational data. The model is further customized through location-based weighting, emphasizing accuracy within a bounding box encompassing the Chilean wind farms, and wind magnitude and power weighting, directly optimizing for these specific metrics. Finally, iterative fine-tuning is combined with linear regression post-processing to further enhance the model’s accuracy. We first fine-tune the base GraphCast model autoregressively on five years (2016-2020) of HRES-fc0 data. During this stage, we used the original GraphCast loss function, a spatially-weighted mean squared error (MSE) calculated across all variables and pressure levels:

\begin{equation}
L_{\text{base}} = \frac{1}{|\mathcal{D}_{\text{batch}}|} \sum_{d_0 \in \mathcal{D}_{\text{batch}}} \frac{1}{T_{\text{train}}} \sum_{\tau \in 1:T_{\text{train}}} \frac{1}{|\mathcal{G}_{0.25^\circ}|} \sum_{i \in \mathcal{G}_{0.25^\circ}} \sum_{j \in J} s_j w_j a_i (\hat x^{d_0+\tau}_{i,j} - x^{d_0+\tau}_{i,j})^2
\end{equation}

\(\mathcal{D}_{\text{batch}}\) denotes a batch of forecast initialization date-times in the training set, \(T_{\text{train}}\) is the number of autoregressive steps used for training, \(\mathcal{G}_{0.25^\circ}\) is the set of grid cells in the 0.25-degree latitude-longitude grid, \(J\) is the set of variables and pressure levels, \(s_j\) represents the inverse variance of time differences for variable \(j\), \(w_j\) is the loss weight for variable \(j\), \(a_i\) is the area of grid cell \(i\), normalized to unit mean over the grid, \(\hat x^{d_0+\tau}_{i,j}\) is the model’s prediction for variable \(j\) at grid cell \(i\) and lead time \(\tau\) from initialization \(d_0\), and \(x^{d_0+\tau}_{i,j}\) is the corresponding target value from the HRES-fc0 data.
We also modify the loss function by incorporating a location-based weighting factor \( m_i \), which up-weights the error contributions from grid cells within a specified bounding box by a factor \( \omega_l \), while grid cells outside the bounding box are left unweighted.

\section{Results}

\textbf{Marginal response of thermal power plants}. By analyzing the coefficients $\beta_1$ and $\beta_2$ of the statistical model, we find that, on average, fluctuations or ramps in thermal generation are more sensitive to increases in wind generation compared to solar. A 1 GWh increase in wind generation displaces 0.95 GWh of thermal generation, whereas the same increase in solar generation displaces only 0.67 GWh, which is 30\% less. We also model the response of individual thermal generators to marginal changes in wind and solar generation by isolating hourly generation in the fixed-effects model. Using these coefficients, we classify each power plant as either ‘solar-following’ or ‘wind-following’ determined based on the absolute value of the larger, statistically significant coefficient. Of the 128 thermal plants, we find that 91 are wind-following, while the remaining 37 largely to solar generation. The vast majority of wind-following thermal plants clearly necessitates better operational wind forecasts in dispatch models used by the ISO.

\begin{figure}[ht!]
    \centering
    \begin{subfigure}{\textwidth}
        \centering
        \includegraphics[width=\textwidth]{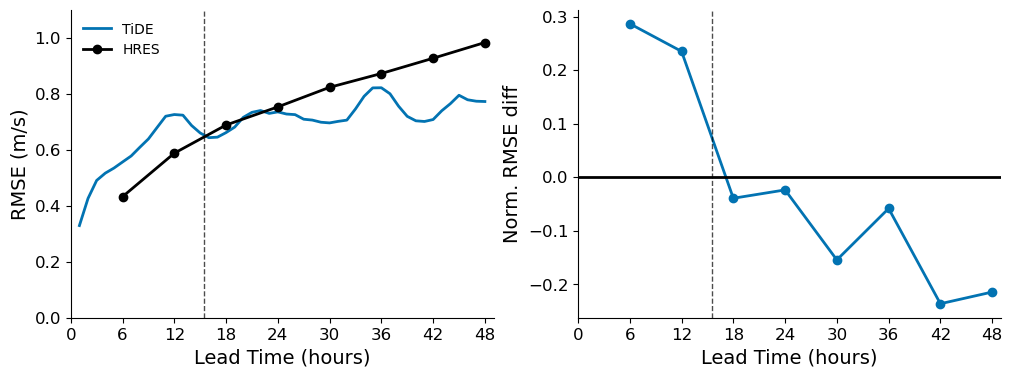}
        \caption{Short-term forecasts (0-2 days)}
        \label{fig:rmse_avg}
    \end{subfigure}
    \vspace{1em}
    \begin{subfigure}{\textwidth}
        \centering
        \includegraphics[width=\textwidth,keepaspectratio]{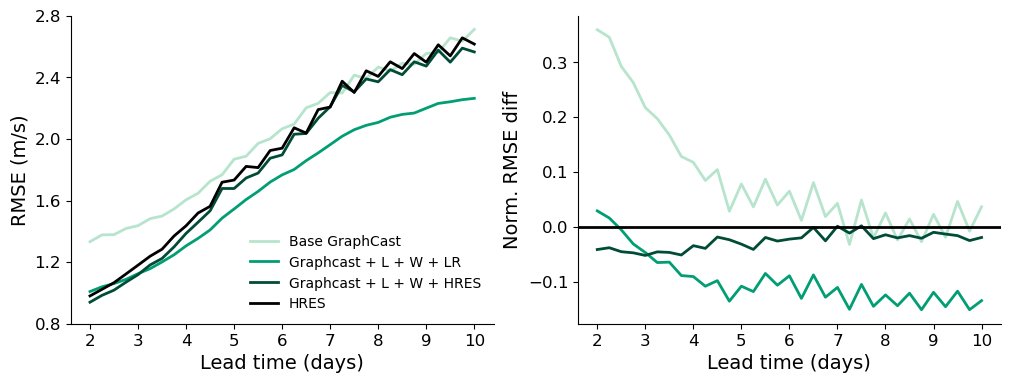}
        \caption{Medium-term forecasts (2-10 days)}
        \label{fig:rmse_GC}
    \end{subfigure}
    \caption{\textbf{Mean hourly RMSE and Normalized RMSE of wind magnitude predictions averaged across all interpolated wind farm locations for 2021}. (a) Short-term forecasts (0-2 days). The left column represents mean hourly wind magnitude RMSE values over a 48-hour forecast horizon, while the right column represents normalized RMSE considering HRES as the baseline using TiDE forecasts at 6-hour intervals. (b) Medium-term forecasts (2-10 days). The left column represents wind magnitude RMSE values over an 8 day forecast horizon at 6-hour time intervals. The solid lines represent three alternate GraphCast models.}
    \label{fig:combined-rmse}
\end{figure}

\textbf{Short-term forecasts}. We evaluate the performance of our TiDE model using the Root Mean Squared Error (RMSE) metric, averaged hourly across all 24 locations, for wind speed. Figure \ref{fig:rmse_avg} shows the metric over a 48-hour horizon for both TiDE and HRES. The RMSE increases gradually over the forecast period for both models; for HRES the data available is at 6-hour intervals. The short-term model performs better than auto-regressive models, as evidenced by a 4-21\% improvement over HRES. The RMSE for HRES forecasts at hours 6 and 12 is lower than that of our short-term model, however, given that operational forecasts are made at 10 am the day prior, the critical forecasting window used for unit commitment and economic dispatch by the ISO is from 14 - 38 hours, during which we consistently outperform HRES. Beyond a 48-hour forecast horizon, TiDE's performance in comparison to HRES resembles a mean-reverting behavior.

\textbf{Medium-term forecasts}. Our custom model weighting training inputs by location, wind magnitude independently, and a combination of both resulted in iterative improvements in skill and RMSE. Figure \ref{fig:rmse_GC} summarizes the RMSE for three fine-tuned GraphCast variants at quarter-degree interpolated wind farm locations. At 96-hour lead times, these weighted models outperform HRES predictions by 10\%, 13\%, and 17\% respectively. The location- and wind-weighted GraphCast has greater wind speed forecasting skill than HRES for periods after a lead time of 30 hours at various pressure levels which roughly correspond to those of wind farm elevations. When further improved by incorporating HRES inputs as a covariate, the crossover point for wind magnitude predictions improves to 2.5 days. Over the medium-term lead times, our finetuned GraphCast model outperforms HRES by 5-23\%.

\section{Acknowledgements}

We thank Andrew El-Kadi, Ferran Alet, and Alexis Boukouvalas for all their guidance with Graphcast. We thank Rodrigo Andres, Leonie Wagner and Farooq Anjum for data management, software, software reviews and experiments, Joshua D'Arcy and Ren Gibbons for early software and experiments on wind prediction, and Abhimanyu Das for guidance in implementing and tuning the TiDE model. We also thank Gautam Bajaj, Sanjana Reddy, and Qianyu Zhang for reviews and comments on the paper. 

\bibliographystyle{plain}
\bibliography{neurips_2023}

\begin{thebibliography}{10}

\bibitem{bi2023accurate}
Kaifeng Bi, Lingxi Xie, Hengheng Zhang, Xin Chen, Xiaotao Gu, and Qi~Tian.
\newblock Accurate medium-range global weather forecasting with 3d neural
  networks.
\newblock {\em Nature}, 619(7970):533--538, 2023.

\bibitem{bushnell2005ownership}
James~B Bushnell and Catherine Wolfram.
\newblock Ownership change, incentives and plant efficiency: The divestiture of
  us electric generation plants.
\newblock 2005.

\bibitem{coordinador}
{Coordinador Eléctrico Nacional}.
\newblock Coordinador eléctrico nacional, 2024.
\newblock Accessed: 2024-05-27.

\bibitem{infotecnica}
{Coordinador Eléctrico Nacional}.
\newblock Información técnica - coordinador eléctrico nacional, 2024.
\newblock Accessed: 2024-05-27.

\bibitem{das2023long}
Abhimanyu Das, Weihao Kong, Andrew Leach, Shaan Mathur, Rajat Sen, and Rose Yu.
\newblock Long-term forecasting with tide: Time-series dense encoder.
\newblock {\em arXiv preprint arXiv:2304.08424}, 2023.

\bibitem{iea_2018}
IEA.
\newblock Energy policies beyond iea countries: Chile 2018 review, 2018.

\bibitem{lam2022graphcast}
Remi Lam, Alvaro Sanchez-Gonzalez, Matthew Willson, Peter Wirnsberger, Meire
  Fortunato, Ferran Alet, Suman Ravuri, Timo Ewalds, Zach Eaton-Rosen, Weihua
  Hu, et~al.
\newblock Graphcast: Learning skillful medium-range global weather forecasting.
\newblock {\em arXiv preprint arXiv:2212.12794}, 2022.

\bibitem{lam2023learning}
Remi Lam, Alvaro Sanchez-Gonzalez, Matthew Willson, Peter Wirnsberger, Meire
  Fortunato, Ferran Alet, Suman Ravuri, Timo Ewalds, Zach Eaton-Rosen, Weihua
  Hu, et~al.
\newblock Learning skillful medium-range global weather forecasting.
\newblock {\em Science}, 382(6677):1416--1421, 2023.

\bibitem{liu2023monitoring}
Zhu Liu, Zhu Deng, Steve Davis, and Philippe Ciais.
\newblock Monitoring global carbon emissions in 2022.
\newblock {\em Nature Reviews Earth \& Environment}, 4(4):205--206, 2023.

\bibitem{molteni1996ecmwf}
Franco Molteni, Roberto Buizza, Tim~N Palmer, and Thomas Petroliagis.
\newblock The ecmwf ensemble prediction system: Methodology and validation.
\newblock {\em Quarterly journal of the royal meteorological society},
  122(529):73--119, 1996.

\bibitem{qiu2022impacts}
Minghao Qiu, Corwin~M Zigler, and Noelle~E Selin.
\newblock Impacts of wind power on air quality, premature mortality, and
  exposure disparities in the united states.
\newblock {\em Science Advances}, 8(48):eabn8762, 2022.

\bibitem{rasp2020weatherbench}
Stephan Rasp, Peter~D Dueben, Sebastian Scher, Jonathan~A Weyn, Soukayna
  Mouatadid, and Nils Thuerey.
\newblock Weatherbench: a benchmark data set for data-driven weather
  forecasting.
\newblock {\em Journal of Advances in Modeling Earth Systems},
  12(11):e2020MS002203, 2020.

\bibitem{suri2024real}
Dhruv Suri, Jacques de~Chalendar, and Ines Azevedo.
\newblock What are the real implications for $ co\_2 $ as generation from
  renewables increases?
\newblock {\em arXiv preprint arXiv:2408.05209}, 2024.

\bibitem{wang2021review}
Yun Wang, Runmin Zou, Fang Liu, Lingjun Zhang, and Qianyi Liu.
\newblock A review of wind speed and wind power forecasting with deep neural
  networks.
\newblock {\em Applied Energy}, 304:117766, 2021.

\end{thebibliography}

\newpage

\appendix

\section{Methods}

\subsection{Statistical Models}

We use a fixed effects ordinary least squares (OLS) regression formulation to model the change in operational behavior of thermal generators in response to a marginal increase in generation from wind and solar. The formulation is a non-logarithmic form of the statistical model used by \cite{bushnell2005ownership} and \cite{suri2024real}.

\begin{equation}
    G_t = \alpha + \beta_1 S_t + \beta_2 W_t + \beta_3 D_t + \beta_4 (W_t - W_{t-1}) + \beta_5 (S_t - S_{t-1}) + \eta_{t,m} + \gamma \textbf{X}_t + \eta_{t,a} + \epsilon
\end{equation}

where $t$ denotes the current timestep; $G_t$, $S_t$, $W_t$ represent the thermal, solar, and wind generation, respectively; $D_t$ is the electricity demand, and the terms between the brackets denote the hourly wind and solar ramp. $\eta_m$ and $\eta_a$ denote the month and year fixed effects. $\textbf{X}_t$ is a set of control variables that includes the hydro and geothermal generation, and imports.

In our formulation, the parameters that are of interest are the coefficients for solar, $\beta_1$ and wind, $\beta_2$. These coefficients represent the change in thermal generation in response to a marginal unit increase in generation from solar and wind, respectively. For example, if generation from wind increases by 1 MWh, then $\beta_1$ and $\beta_2$ will represent the MWh change in generation from thermal plants in Chile. These coefficients thus measure the causal change in the dependent variable under the assumption that daily wind and solar production (represented by $W$ and $S$) is uncorrelated with the error term $\epsilon$ after controlling for demand  and time fixed effects. As suggested by \cite{qiu2022impacts}, this assumption is valid for wind at the hourly level and can thus be extended into the present form of the statistical model.

\subsection{Short-term Wind Forecasting}

The Time-series Dense Encoder (TiDE) model represents a state-of-the-art approach for short-term wind speed forecasting. This model employs a Multi-layer Perceptron (MLP) based encoder-decoder architecture, optimized to handle time-series data with multiple covariates and complex, non-linear dependencies. The encoder processes historical wind speed data and relevant covariates, transforming them into a latent space representation through multiple residual blocks, which are crucial for capturing intricate patterns in the data. The decoder then reconstructs future wind speeds from this latent representation, allowing the model to make accurate predictions up to 48 hours ahead.

The hyperparameters of the TiDE model are chosen to enhance its predictive performance, tuned on the validation loss for each interpolated wind-farm location. Ranges of different hyperparameters are shown in Table \ref{tab:parameter_ranges}. Key hyperparameters include the number of layers in the encoder and decoder, the hidden layer sizes, and dropout rates to prevent overfitting. Specifically, the encoder and decoder each consist of four residual blocks, with each block containing fully connected layers of 128 neurons. Dropout is applied with a rate of 0.1 to ensure regularization. The model is trained using the Mean Squared Error (MSE) loss function, which measures the average squared difference between predicted and actual wind speeds. The Adam optimizer, with a learning rate of 0.001, is utilized to update the model's weights efficiently during training. An early stopping mechanism with a patience of 20 epochs is implemented to halt training when the validation loss ceases to improve, thus avoiding overfitting.

\begin{table}[h!]
    \centering
    \caption{Ranges of different hyperparameters for model tuning}
    \begin{tabular}{l l}
        \toprule
        \textbf{Parameter} & \textbf{Range Evaluated} \\
        \midrule
        Hidden Size            & [256, 512, 1024] \\
        Number of Encoder Layers & [1, 2, 3] \\
        Number of Decoder Layers & [1, 2, 3] \\
        Decoder Output Dimension & [4, 8, 16, 32] \\
        Temporal Decoder Hidden  & [32, 64, 128] \\
        Dropout Level           & [0.0, 0.1, 0.2, 0.3, 0.5] \\
        Layer Normalization     & [True, False] \\
        Learning Rate           & Log-scale in [1e-5, 1e-2] \\
        Reversible Instance Norm (RevIN) & [True, False] \\
        \bottomrule
    \end{tabular}
    \label{tab:parameter_ranges}
\end{table}

The TiDE model's performance is evaluated using the Root Mean Squared Error (RMSE) metric, averaged hourly across multiple locations in Chile. To handle the time-series nature of the data, the model incorporates various datetime attributes as additional covariates, such as the hour of the day, day of the week, and month, which are encoded using dense layers in a Multi-Layer Perceptron (MLP)-based architecture. Additionally, the model is equipped to use a Scaler transformer to normalize the input data, ensuring that all features have zero-mean unit variance. This preprocessing step is essential for maintaining the stability and performance of the neural network. We normalize and detect outliers using a z-scoring threshold specified in \cite{bi2023accurate}. By adopting a higher z-score threshold, performance at later lead times beyond 24 hours in not affected. The inclusion of future covariates, such as temperature, pressure, and various wind parameters at different altitudes, enhances the model's ability to capture the underlying weather patterns, leading to more accurate short-term wind speed forecasts.

\subsubsection{Integration with Randomized Iterative Forecasting}
To further enhance the predictive performance and computational efficiency of the TiDE model, we integrate a randomized iterative forecasting framework into the model. This integration is inspired by the decomposition-ensemble learning methodology where the data is first decomposed into simpler components, which are then modeled individually used to condition the TiDE model on randomly selected time intervals during training. At inference time, the model uses the learned intervals to generate multiple forecasts for a specific lead time. For example, to predict the weather 7 days ahead, the model might use a combination of forecasts at 12-hour intervals repeated 14 times or at 24-hour intervals repeated 7 times. These forecasts are then averaged to produce the final prediction, significantly improving accuracy by reducing the uncertainty inherent in long-term forecasts.

\subsubsection{Model Architecture and Mathematical Formulation}

Let $\mathbf{y}_{1:L}$ represent the historical wind speeds, $\mathbf{x}_{1:L+H}$ the covariates over the look-back and forecast horizon, and $\mathbf{a}$ the static attributes. The encoding step maps the past covariates and wind speeds into a dense representation $\mathbf{e}$:

\begin{equation}
\mathbf{x}_{t} = \text{ResidualBlock}(\mathbf{x}_{t})
\end{equation}

\begin{equation}
\mathbf{e} = \text{Encoder}(\mathbf{y}_{1:L}; \mathbf{x}_{1:L+H}; \mathbf{a})
\end{equation}

This encoding $\mathbf{e}$ is then utilized by the decoder to predict future wind speeds.

The decoding process in the TiDE model involves mapping the encoded hidden representations into future predictions. The dense decoder processes the encoding $\mathbf{e}$ to produce an intermediate representation, which is further refined by the temporal decoder to generate the final forecast. The process is described by the following equations:

\begin{equation}
\mathbf{g} = \text{Decoder}(\mathbf{e})
\end{equation}

\begin{equation}
\mathbf{D} = \text{Reshape}(\mathbf{g})
\end{equation}

\begin{equation}
\hat{\mathbf{y}}_{L+t} = \text{TemporalDecoder}(\mathbf{d}_{t}; \mathbf{x}_{L+t})
\end{equation}

where $\mathbf{g}$ is the output of the dense decoder, $\mathbf{D}$ is the reshaped matrix of decoded vectors, and $\mathbf{d}_{t}$ is the decoded vector for time $t$. The temporal decoder combines this decoded vector with the projected covariates $\mathbf{x}_{L+t}$ to generate the final prediction $\hat{\mathbf{y}}_{L+t}$.

\subsubsection{Randomized Algorithm-based Decomposition-Ensemble Learning}

The proposed method involves three major steps: data decomposition, individual prediction using randomized algorithms, and results ensemble.

Data Decomposition: We employ an effective multi-scale analysis technique such as Ensemble Empirical Mode Decomposition (EEMD) to decompose the original time series data into $n$ intrinsic mode functions (IMFs) and a residue. This decomposition ensures that the data complexity is reduced, making it easier to model.

\begin{equation} 
\mathbf{x}_t = \sum_{j=1}^{n} \mathbf{c}_{j,t} + \mathbf{r}_{t} 
\end{equation}

Individual Prediction using Randomized Algorithms: For each decomposed component, instead of forecasting weather conditions at fixed intervals, we condition the TiDE model on randomly selected time intervals during training. This is done by introducing a distribution over time intervals, choosing from intervals such as 6, 12, 24, and 48 hours. This randomization serves a dual purpose: it acts as a form of data augmentation by varying the intervals during training, and it enables the model to produce multiple forecasts for a specified lead time during inference by combining predictions from different intervals.

Results Ensemble: The final forecast is obtained by aggregating the predictions of all decomposed components and the residue.

\begin{equation} 
\mathbf{\hat{x}}_{t} = \sum_{j=1}^{n} \mathbf{\hat{c}}_{j,t} + \mathbf{\hat{r}}_{t} 
\end{equation}

\subsubsection{Performance Evaluation}

Mathematically, the forecasting function $f$ can be defined as:

\begin{equation}
f : \left( \left\{ \mathbf{y}_{1:L}^{(i)} \right\}_{i=1}^{N}, \left\{ \mathbf{x}_{1:L+H}^{(i)} \right\}_{i=1}^{N}, \left\{ \mathbf{a}^{(i)} \right\}_{i=1}^{N} \right) \rightarrow \left\{ \sum_{j=1}^{n} \hat{\mathbf{c}}_{L+1:L+H,j}^{(i)} + \hat{\mathbf{r}}_{L+1:L+H}^{(i)} \right\}_{i=1}^{N}
\end{equation}

where:
\begin{itemize}
    \item $N$ is the number of time-series,
    \item $L$ is the look-back period,
    \item $H$ is the forecast horizon,
    \item $\hat{\mathbf{c}}_{L+1:L+H,j}^{(i)}$ represents the predicted values for the $j$-th IMF,
    \item $\hat{\mathbf{r}}_{L+1:L+H}^{(i)}$ represents the predicted residue.
\end{itemize}

The accuracy of the prediction is measured by the Root Mean Squared Error (RMSE), defined as:

\begin{align}
\text{RMSE}\left( \left\{ \mathbf{y}_{L+1:L+H}^{(i)} \right\}_{i=1}^{N}, \left\{ \sum_{j=1}^{n} \hat{\mathbf{c}}_{L+1:L+H,j}^{(i)} + \hat{\mathbf{r}}_{L+1:L+H}^{(i)} \right\}_{i=1}^{N} \right) = \nonumber \\
\sqrt{\frac{1}{NH} \sum_{i=1}^{N} \left\| \mathbf{y}_{L+1:L+H}^{(i)} - \left( \sum_{j=1}^{n} \hat{\mathbf{c}}_{L+1:L+H,j}^{(i)} + \hat{\mathbf{r}}_{L+1:L+H}^{(i)} \right) \right\|_2^2}
\end{align}

This integration of randomized iterative forecasting with TiDE leverages the speed and robustness of randomized algorithms to enhance the efficiency and accuracy of wind speed predictions, particularly in scenarios involving large-scale time series data with complex patterns.

\subsection{GraphCast}

GraphCast is a neural network architecture designed to predict weather states by taking the two most recent states of Earth's weather - the current time and six hours earlier - and forecasting the next state six hours ahead. The model uses a 0.25-degree latitude-longitude grid to represent a single weather state, corresponding to roughly 28 km by 28 km resolution at the equator. Implemented as a neural network architecture based on Graph Neural Networks (GNNs) in an "encode-process-decode" configuration, GraphCast consists of 36.7 million parameters.

The encoder within GraphCast employs a single GNN to map variables, normalized to zero-mean unit variance, represented as node attributes on the input grid to learned node attributes on an internal multi-mesh representation. This multi-mesh is spatially homogenous and defined by refining a regular icosahedron iteratively six times, leading to a high-resolution graph with 40,962 nodes and a flat hierarchy of edges of varying lengths. The processor then uses 16 unshared GNN layers to perform learned message-passing on the multi-mesh, enabling efficient local and long-range information propagation with minimal message-passing steps. The decoder maps the final processor layer’s learned features back to the latitude-longitude grid using a single GNN layer, predicting the output as a residual update to the most recent input state, with output normalization to achieve unit-variance on the target residual.

For fine-tuning GraphCast to optimize wind magnitude and power forecasts specifically for Chilean wind farms, a multi-stage fine-tuning process was employed. This involved autoregressive fine-tuning on HRES-fc0 data to account for potential distribution shifts between reanalysis data (ERA5) and real-time operational data. The model was further customized through location-based weighting, emphasizing accuracy within a bounding box encompassing the Chilean wind farms, and wind magnitude and power weighting, directly optimizing for these specific metrics. Iterative fine-tuning combined with linear regression post-processing enhanced the model’s accuracy, making it a robust tool for predicting wind patterns in Chile’s diverse and complex terrain.

\section{Finetuning GraphCast}

GraphCast demonstrates strong global weather forecasting skills, but our objective is to optimize wind magnitude and power predictions specifically for Chilean wind farms. To achieve this, we customized the publicly available GraphCast model, pre-trained on ERA5 reanalysis data, using a multi-stage fine-tuning process. This section elaborates on each stage and provides the mathematical formulation of the modified loss function.

\subsection{Operational Data Fine-tuning}

Recognizing the potential distribution shift between reanalysis data (ERA5) and real-time operational data (HRES-fc0), we first fine-tuned the base GraphCast model autoregressively on five years (2016-2020) of HRES-fc0 data. This fine-tuning involved several steps. Firstly, HRES-fc0 data was pre-processed to align with the GraphCast input format. This included variable selection, focusing on relevant variables directly related to wind dynamics, such as eastward and northward wind components at various pressure levels. Additionally, the data was converted to the same spatial resolution as GraphCast (0.25-degree latitude-longitude grid).

The autoregressive fine-tuning process involved 30,000 steps where the model predicted the weather state six hours ahead for up to a ten-day lead time. In each training step, the model’s output from the previous timestep was fed back as input, simulating a real-time forecasting scenario. This recursive process allowed the model to learn temporal dependencies and adapt to the dynamics of operational data. During this stage, we used the original GraphCast loss function, a spatially-weighted mean squared error (MSE) calculated across all variables and pressure levels:

\begin{equation}
L_{\text{base}} = \frac{1}{|\mathcal{D}_{\text{batch}}|} \sum_{d_0 \in \mathcal{D}_{\text{batch}}} \frac{1}{T_{\text{train}}} \sum_{\tau \in 1:T_{\text{train}}} \frac{1}{|\mathcal{G}_{0.25^\circ}|} \sum_{i \in \mathcal{G}_{0.25^\circ}} \sum_{j \in J} s_j w_j a_i (\hat x_{i,j,d_0+\tau} - x_{i,j,d_0+\tau})^2
\end{equation}

In this equation, \(\mathcal{D}_{\text{batch}}\) denotes a batch of forecast initialization date-times in the training set, \(T_{\text{train}}\) is the number of autoregressive steps used for training, \(\mathcal{G}_{0.25^\circ}\) is the set of grid cells in the 0.25-degree latitude-longitude grid, \(J\) is the set of variables and pressure levels, \(s_j\) represents the inverse variance of time differences for variable \(j\), \(w_j\) is the loss weight for variable \(j\), \(a_i\) is the area of grid cell \(i\), normalized to unit mean over the grid, \(\hat x_{i,j,d_0+\tau}\) is the model’s prediction for variable \(j\) at grid cell \(i\) and lead time \(\tau\) from initialization \(d_0\), and \(x_{i,j,d_0+\tau}\) is the corresponding target value from the HRES-fc0 data.

\subsection{Location and Wind-Magnitude Weighting}

To focus the model’s predictive skill on Chilean wind farms, we applied location-based weighting and wind magnitude weighting. The location-based weighting emphasized the accuracy within a bounding box encompassing the Chilean wind farms, while the wind magnitude weighting directly optimized for predicting wind weather variables. This involved adjusting the weights in the loss function to prioritize errors in these regions and variables, thereby fine-tuning the model to the specific operational characteristics of the Chilean wind farms.

The location-based weighting modified the loss function as follows:

\begin{equation}
L_{\text{loc}} = \frac{1}{|\mathcal{D}_{\text{batch}}|} \sum_{d_0 \in \mathcal{D}_{\text{batch}}} \frac{1}{T_{\text{train}}} \sum_{\tau \in 1:T_{\text{train}}} \frac{1}{|\mathcal{G}_{0.25^\circ}|} \sum_{i \in \mathcal{G}_{0.25^\circ}} \sum_{j \in J} s_j w_j a_i m_i (\hat x_{i,j,d_0+\tau} - x_{i,j,d_0+\tau})^2
\end{equation}

where \(m_i\) is defined as:

\begin{equation}
m_i =
\begin{cases} 
\omega_l, & \text{if grid cell } i \text{ is within the bounding box} \\
1, & \text{otherwise}
\end{cases}
\end{equation}

and \(\omega_l\) is the location-based up-weighting factor.

For wind magnitude and power weighting, we introduced additional loss terms to optimize the model for predicting wind magnitude (\(wm\)) and power (\(wp\)). Wind magnitude was calculated from the predicted eastward (\(u_{10m}\)) and northward (\(v_{10m}\)) wind components at 10 meters:

\[
wm = \sqrt{u_{10m}^2 + v_{10m}^2}
\]

Wind power was estimated using a simplified cubic relationship:

\[
wp = wm^3
\]

\subsection{Iterative Fine-tuning and Configuration Selection}

An iterative fine-tuning process was employed to refine the model further. Various configurations were tested, adjusting hyperparameters and re-evaluating performance to identify the optimal setup. The configuration achieving the lowest RMSE for Chilean wind farm locations was chosen as the final fine-tuned model.

\subsection{Linear Regression Post-processing}

To enhance accuracy, we applied a per-lead-time bias correction using linear regression. This involved training separate linear regression models for each six hour lead time (up to 10 days), variable (e.g., \(u_{10m}\), \(v_{10m}\), wind magnitude, and power), and grid cell within the bounding box. The last year (2020) of HRES-fc0 data was used for training these models, with the results evaluated on 2021. The correction formula is given by:

\begin{equation}
\hat{y}_{\text{corrected}} = \alpha + \beta \hat{y}_{\text{raw}}
\end{equation}

where \(\hat{y}_{\text{corrected}}\) is the bias-corrected prediction, \(\hat{y}_{\text{raw}}\) is the raw GraphCast prediction, and \(\alpha\) and \(\beta\) are the learned intercept and slope, respectively. These models were applied to the fine-tuned GraphCast predictions to generate the final bias-corrected wind forecasts.

\subsection{HRES Initialized Inputs}

To further enhance the predictive accuracy of our location and wind-magnitude weighted GraphCast model, we investigated the integration of real-time forecasts from the ECMWF’s High-Resolution Forecasting System (HRES) as additional input features. Building on existing work in this field, our "HRES forcing" approach leverages the inherent short-term predictive skill of HRES to potentially inform and improve the medium-range wind speed forecasts generated by the GraphCast model.

Implementation of HRES forcing involves augmenting the input layer of our customized GraphCast model to accommodate the HRES forecast variables. The weights associated with these new inputs are initialized as zeros, preserving the model’s pre-existing functionality, which is derived from pre-training on ERA5 and operational fine-tuning on historical HRES-fc0 data. A subsequent single-step autoregressive fine-tuning process, utilizing a reduced number of steps (2,000 compared to the standard 30,000), allows the model to learn how to effectively incorporate the real-time HRES information while minimizing disruption to the previously established, regionally-tailored weights.

\begin{figure}[h]
    \centering
    \includegraphics[width=\textwidth]{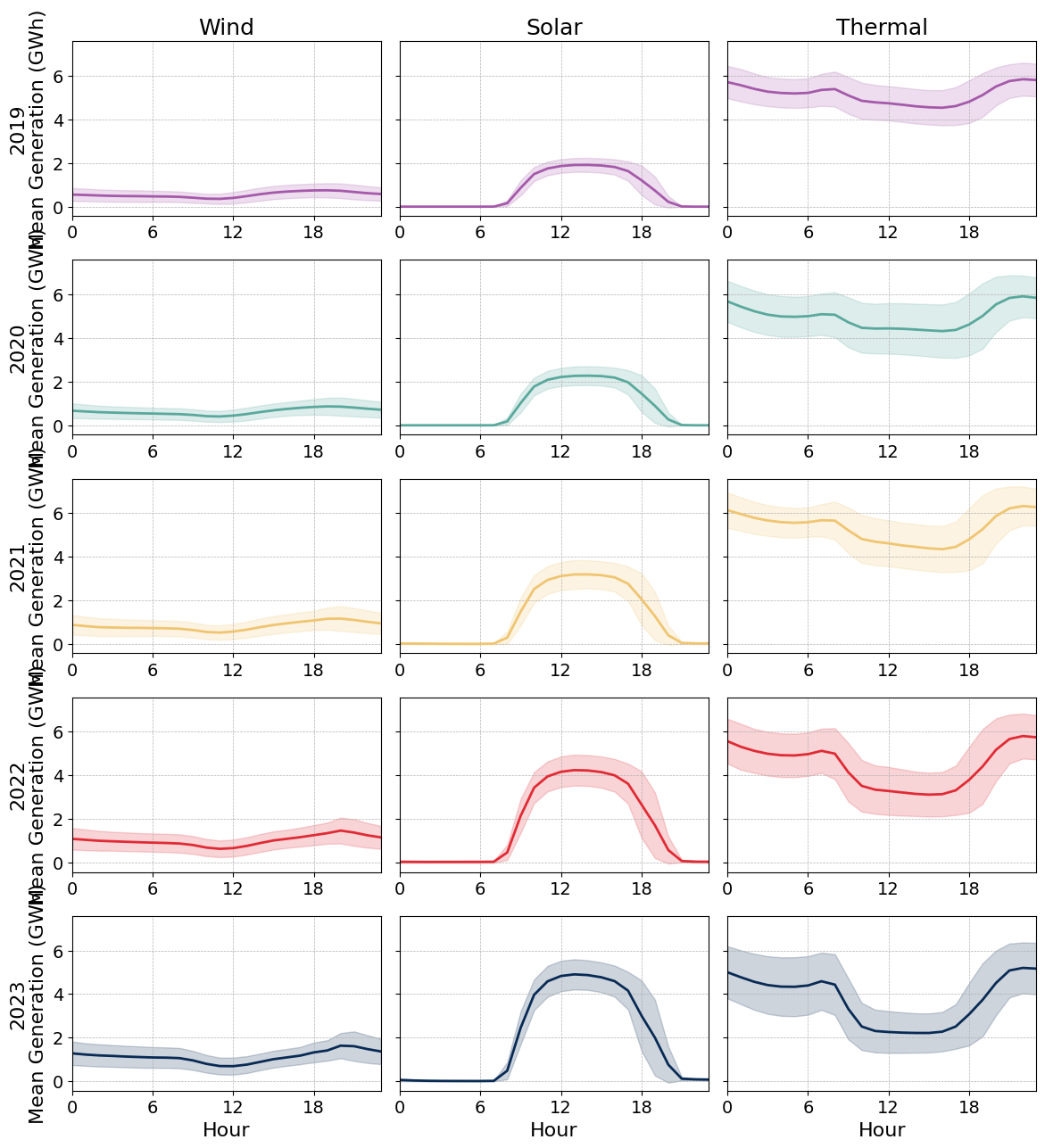}
    \caption{\textbf{Mean hourly generation and standard deviation for wind, solar, and thermal power plants in Chile from 2019 to 2023}. Each row corresponds to a different year, and each column represents a different technology (Wind, Solar, and Thermal). The solid lines indicate the mean generation in gigawatt-hours (GWh), and the shaded areas represent the standard deviation across all hours. The x-axis denotes the hour of the day, ranging from 0 to 23. This figure illustrates the temporal variation in power generation for each technology over the years, highlighting both the average generation and the variability within each day.
}
    \label{fig:Figure_1}
\end{figure}

\begin{figure}[h]
    \centering
    \includegraphics[width=\textwidth]{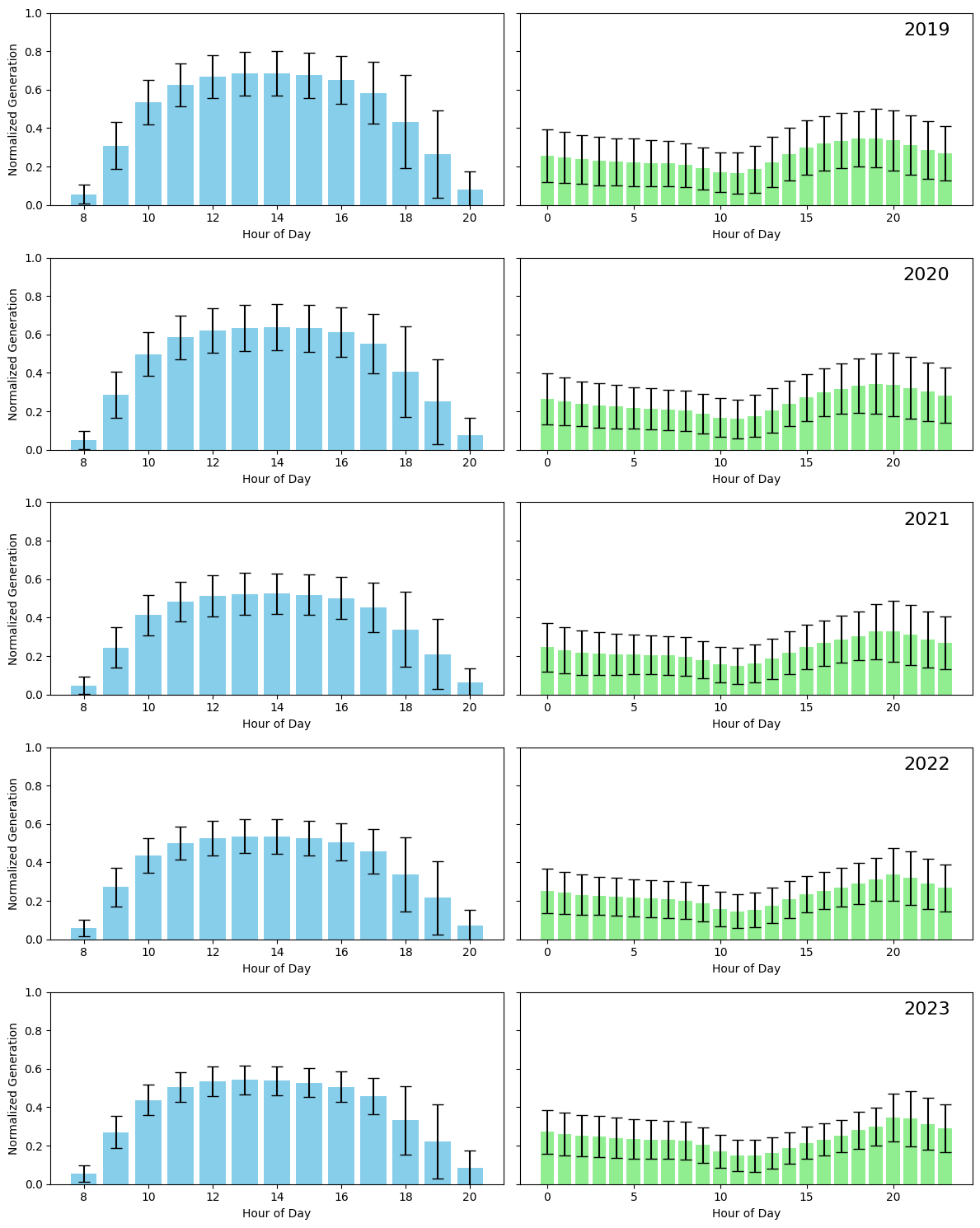}
    \caption{\textbf{Hourly solar and wind generation (2019-2023)}. Solar (left) and Wind (right). The colored bars represent mean generation normalized by installed capacity by source and year. The bars indicate the standard deviation of hourly normalized generation by source.
}
    \label{fig:Figure_S1}
\end{figure}

\begin{figure}[ht!]
    \centering
    \includegraphics[width=0.9\textwidth]{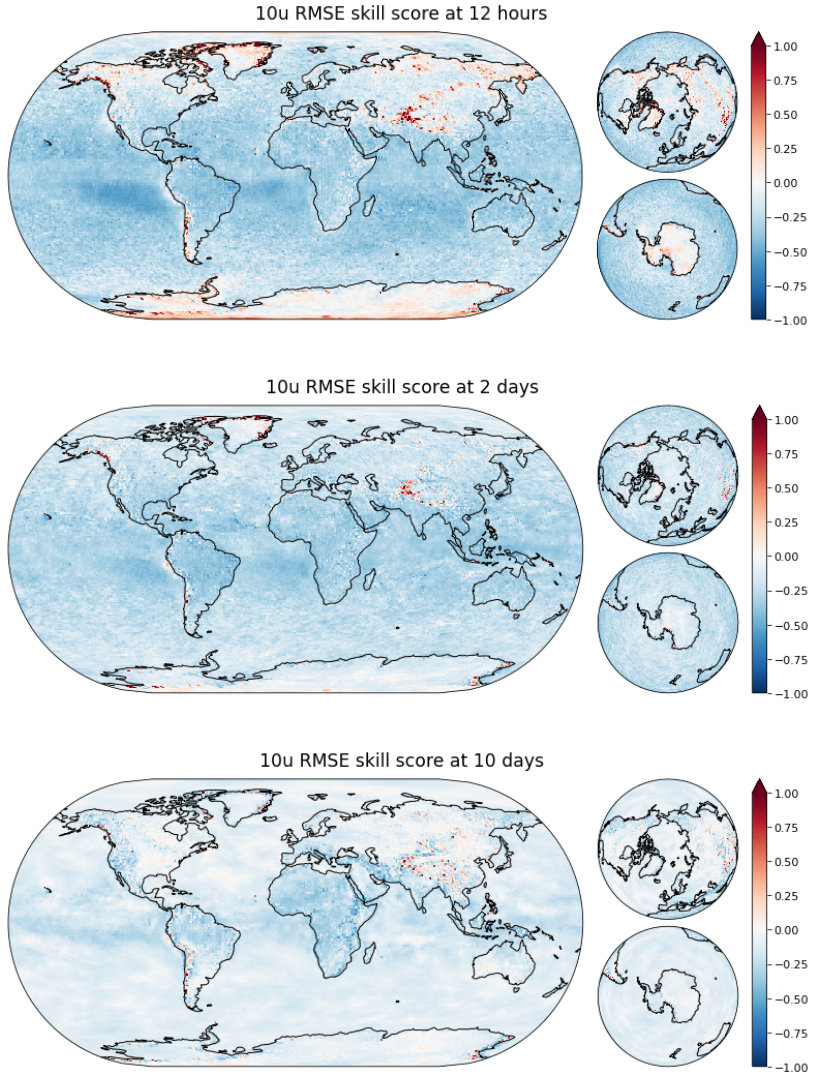}
    \caption{\textbf{Normalized RMSE difference of GraphCast's 10u forecasts relative to HRES, by location, at 12 hours, 2 days and 10 day lead times}. Blue indicates that GraphCast has greater skill than HRES, Red that HRES has greater skill. Here "10u" refers to the u-component of wind at an altiitude corresponding to 10m \cite{lam2022graphcast}}
    \label{fig:GC_RMSE}
\end{figure}

\end{document}